# A Feed-Forward Artificial Intelligence Pipeline for Sustainable Desalination under Climate Uncertainties: UAE Insights

Obumneme Nwafor[a] [*], Chioma Nwafor[b], Amro Zakaria[c], Nkechi Nwankwo[d]

[a] *School of Computing, Engineering and Built Environment, Glasgow Caledonian University, Scotland*
obumneme.nwafor@gcu.ac.uk  ORCID: 0000-0002-0993-1659

[b] *Glasgow School of Business and Society , Glasgow Caledonian University, Scotland, Email:* chioma.nwafor@gcu.ac.uk
ORCID: 0000-0001-9612-7214

[c]*Kyoto Network, Abu Dhabi, United Arab Emirates,* Amro@kyoto.network ORCID: 0009-0003-4192-6950

[d] *Glaggow Caledonian University, London Campus,* Nkechi.nwankwo@gcu.ac.uk  ORCID: 0009-0004-6769-0031

[*] **CORRESPONDING AUTHOR:** Obumneme Nwafor, obumneme.nwafor@gcu.ac.uk

**Abstract**

The United Arab Emirates (UAE) relies heavily on seawater desalination to meet over 90% of its drinking water needs. Desalination processes are highly energy-intensive, accounting for approximately 15% of the UAE's electricity consumption and contributing to over 22% of the country's energy-related $CO_2$ emissions. More so, theses processes face significant sustainability challenges in the face of climate uncertainties such as rising seawater temperatures, salinity, and aerosol optical depth (AOD). AOD greatly affects the operational and economic performance of solar-powered desalination systems through photovoltaic soiling, membrane fouling and water turbidity cycles. This study proposes a novel pipelined two-stage predictive modelling architecture: the first stage forecasts AOD using satellite-derived time series and meteorological data; the second stage uses the predicted AOD and other meteorological factors to predict desalination performance efficiency losses. The framework achieved 98% accuracy and SHapley Additive exPlanations (SHAP) explainability was used to reveal key drivers of the system degradation. Furthermore, this study proposes a dust-aware rule-based control logic for desalination systems based on predicted values of AOD and solar efficiency. This control logic is used to adjust the desalination plant feed water pressure, adapt maintenance scheduling logic and regulate energy source switching. To enhance practical utility of the research findings, the predictive models and rule-based controls were packaged into an interactive dashboard for scenario and predictive analytics. This provides a management decision-support system for climate-adaptive planning.

**Keywords:** Artificial Intelligence, Desalination, UAE, Satellite Forecasting, AOD

## 1. Introduction

Seawater desalination plays a very important role in ensuring access to safe drinking water in arid regions like the UAE where the country is pioneering energy optimisation strategies in desalination plants. This includes the adoption of advanced desalination technologies and integration of solar and other renewable energy sources to enhance efficiency and sustainability (Al-Addous et al., 2024; Khuwaileh et al., 2023). However, AOD plays a major role in affecting the operational performance of solar-powered desalination systems in arid regions where the water-energy dynamics is particularly sensitive to environmental extremes. Dust intrusion affects photovoltaic (PV)-driven desalination facilities through multiple mechanisms such as photovoltaic soiling (Ammari et al., 2022), membrane fouling (Ahmed et al., 2023; Valdés et al., 2021), and turbidity (Al Kaabi et al., 2012). These physical mechanisms translate into serious economic impacts on desalination plants. For instance, increased PV cleaning frequency costs approximately $0.35 per square meter per session (Tabassum and Rizwan, 2020), membrane replacements incur up to $200,000 annually per plant and energy penalties during dust events can reach 4 kWh/m² for a 10 MW solar facility which is approximately $12–$20/MWh based on typical UAE electricity prices of $0.03–$0.05 per kWh) (Kosmopoulos et al., 2018). Current approaches to optimising desalination mostly rely on static models without dynamic adaptation to environmental volatility. This study introduces a hybrid approach that leverages satellite-based AI prediction and rule-based control logic to optimise desalination plant efficiency.

## 2. Literature Review and Theoretical Background

A significant body of research has applied AI and Computational Techniques to study Energy Optimisation, Renewable Energy Integration and Sustainability in UAE desalination systems. For instance, model-based optimisation applied to the Arab Potash Company plant achieved a 27.97% energy reduction by tuning inlet conditions such as feed pressure, flow rate, and temperature (Mudhar et al, 2023). Similar work integrating microbial desalination cells (MDCs) as a pre-treatment to revers osmosis (RO) has shown substantial reductions in energy demands while simultaneously addressing wastewater treatment and electricity generation (Tarun & Vahid, 2021). The integration of renewable energy sources (RES) into desalination has also been widely investigated: a hybrid photovoltaic/thermal (PVT) system achieved 70% renewable energy utilisation and an 18% reduction in water cost (Alqaed et al., 2021). Another study modelled cogeneration and solar-based multi-generation systems that include desalination, cooling, and hydrogen production, showing performance gains using Grey Wolf optimization algorithms (Navid et al., 2022). These findings highlight the role of hybrid and multi-generational configurations in maximizing energy utility and resilience in the UAE and other Gulf states with similar climate conditions. AI models such as Grey Wolf Optimiser-Artificial Neural Network (GWO-ANN) model achieved a predictive accuracy of R² = 98.9%, enabling enhanced decision-making in plant operation (Mahadeva et al, 2022). Likewise, a Genetic Neuro-Fuzzy system demonstrated success in dynamically adjusting pump operations for on-board SWRO systems, potentially transferable to land-based units (López et al., 2023). Also, a demand-side management models leveraging Time-of-Use (TOU) pricing with RES and diesel backup achieved energy savings and freshwater production increases (Okampo & Nwulu, 2020). Smart-grid cooperation models were used to reduce lifetime desalination costs by 60% through optimal sizing and energy exchange (Malisovas & Koutroulis, 2020). Despite the UAE's vulnerability to climate uncertainty, there is still limited research that incorporate stochastic climate data into desalination planning (Ahamad et al., 2023).

## 2.1 Solar Irradiance and Efficiency

Solar irradiance plays a major role in energy input estimation solar-powered desalination plants (Al-Obaidi et al., 2022; Mundu et al., 2025). Solar irradiance is the power per unit area received from the sun in the form of electromagnetic radiation (Chu et al., 2023). It is typically expressed in Watts per square meter (W/m²). There are two forms of irradiance used in this study: the Clear-Sky Irradiance, which represents the theoretical solar radiation received at ground level assuming no clouds or aerosols (Paulescu & Paulescu, 2021). It serves as a benchmark for maximum solar potential. The second is the Actual Irradiance, which is a measure of the solar radiation at the site, which is often reduced by clouds, dust, or pollution (Zhu et al., 2025). The efficiency loss due to atmospheric effects can be quantified as:

$$Efficiency\ loss\ (\%) = 100 \times \frac{I_{clear} - I_{actual}}{I_{clear}} \quad (1)$$

Equation (1) reflects the reduction in available solar energy, which directly impacts the energy input for desalination processes and integrated photovoltaic performance.

## 2.2 Aerosol Optical Depth (AOD)

AOD is a dimensionless measure of the reduction of solar radiation by aerosol particles in the atmosphere (Torres & Fuertes 2021). It quantifies the level of aerosol loading such as dust, smoke, and pollution. An AOD of 0.0 indicates a crystal-clear sky, while values above 0.4 indicate high aerosol concentration. Higher AOD values mean more scattering and absorption of sunlight before it reaches the Earth's surface, which results in lower actual irradiance and consequently reduced efficiency in solar energy systems that power or support desalination plants. AOD is typically retrieved from satellite measurements instruments such as MODIS (Moderate Resolution Imaging Spectroradiometer) which provide daily AOD data at approximately 1 km resolution (Xinghong et al, 2022). Studies such as Aloïs et al (2021) and Akriti et al. (2021) have shown strong inverse relationships between AOD and photovoltaic efficiency, highlighting its significance in energy forecasting. AOD to solar irradiance attenuation can be expressed using the modified Beer-Lambert Law for dust (Yan et al., 2022) as follow:

$$I_{actual} = I_{clear-sky} \cdot e^{-AOD \cdot m} \quad (2)$$

Where $I_{actual}$ is the observed irradiance (W/m²), $I_{clear-sky}$ is the theoretical maximum irradiance (from NASA POWER), AOD is the Aerosol Optical Depth at 550nm (from MODIS) and $M$ is the air mass coefficient (1/cos(θ), θ = solar zenith angle). Equation (2) show that higher AOD (more aerosols) results in lower surface solar irradiance, which in turn reduces efficiency as expressed in equation(1). These changes in irradiance (affected by AOD) can impact freshwater production efficiency in solar thermal desalination system, where the overall efficiency of the system can be approximated by:

$$\eta_{ther} = \frac{m \times h_{fg}}{A \times I} \quad (3)$$

Where m is the mass flow rate of freshwater (kg/s), $h_{fg}$ is the latent heat of vaporization (kJ/kg), $A$ represents the collector surface area (m²) and I is the incident solar irradiance (W/m²). This impact is significant because photovoltaic (PV) efficiency depends on irradiance as expressed by equation (4) below:

$$\eta_{pv} = \eta_{ref} \times (1 + \alpha(I - I_{ref})) \qquad (4)$$

Where $\eta_{pv}$ is the actual PV efficiency, $\eta_{ref}$ represents the reference PV efficiency, α is the irradiance sensitivity coefficient, $I$ is the current irradiance and $I_{ref}$ is the reference irradiance.

## 2.3 Current Research Gap and Contributions

AOD is widely acknowledged as an important atmospheric factor that modulates solar irradiance, thereby directly influencing the efficiency of solar-powered energy systems. While previous studies have explored AOD dynamics for climatological monitoring and air quality assessment (Mohammadpour et al., 2022; Rahman et al., 2024), a significant gap still exists in applying AOD forecasting within the operational context of solar desalination systems arid regions. Most existing approaches treat AOD prediction and desalination performance as independent and disjointed problems, therefore there is no integrated coherent decision-support framework. This disjointed approach presents a serious limitation because the real-time influence of AOD (atmospheric dust) on solar irradiance, and consequently on the operational energy demands of desalination systems, remains under-researched. Many current desalination optimization models rely on historical or static irradiance profiles, which fails to capture short-term variability introduced by aerosol concentrations, despite evidence that such variability can lead to significant deviations in system efficiency (Feron et al., 2021; Cai et al., 2021).

To address these gaps, this study proposes a novel two-stage end-to-end AI prediction pipeline that first forecasts AOD using deep learning models, and then dynamically feeds the predicted AOD values into a subsequent energy efficiency prediction model for solar-powered desalination systems. Using predicted AOD instead of historical AOD offers a proactive, dynamic and prescriptive approach instead of reactive static and descriptive. Furthermore, while historical AOD reflects past dust conditions, predicted AOD allows for optimisation for future conditions based on expected interference and scenario simulation for risk management and real-time or day-ahead planning. This sequential architecture reflects the causal chain from atmospheric aerosols to system performance which forms an end-to-end system that emulates the real-world forecasting-to-operations flow. More so, this study contributes to the discuss on the use of explainability AI to enhance transparent insights, trust and auditability of predictive models (Nwafor et al., 2023, 2022; Hassija et al., 2024).

## 3. Methodology
### 3.1 Dataset
The dataset for this study was curated by integrating data from satellite sources using a pipeline built with Google Earth Engine (GEE) and Python. First, the Region of Interest (ROI) was defined, a geographic bounding polygon covering the United Arab Emirates. Then, daily meteorological data was obtained from NASA's Prediction of Worldwide Energy Resources (POWER) project. Subsequently, AOD measurements were retrieved from the Moderate Resolution Imaging Spectroradiometer (MODIS) collection, filtered to include only the 0.47μm wavelength band most relevant for aerosol monitoring. For each date in the NASA dataset, the corresponding daily AOD was computed by averaging pixel values around the Dubai Solar Park. These AOD values were then merged with the NASA POWER data based on matching dates. Missing AOD entries were interpolated, and a new feature, *efficiency loss* *percentage*, was computed as the relative difference between clear-sky and actual irradiance. The resulting dataset consists of daily atmospheric aerosol levels, solar irradiance, temperature, dew point, wind speed, humidity, and surface pressure, for UAE from January 2010 to December 2023.

The methodology used in this study integrates time-series analysis, advanced regression techniques, and neural networks to predict both AOD and solar efficiency loss. The first stage of the framework involves predicting AOD values using hybrid Bi-LSTM and XGBoost models. Considering the effect of both static and temporal dynamics of AOD is very important in order to accurately forecast AOD and understand solar attenuation in dusty environment. The static meteorological features (X_static) used in the model include daily surface temperature (T2M), dew point (T2MDEW), wind speed (WS2M), humidity (QV2M), surface pressure (PS), and the month of the year. These variables are important determinants of atmospheric behaviour that influence aerosol formation, dispersion, and deposition based on previous studies in the literature review section. The temporal sequence features (X_seq) are engineered from historical AOD values. The first temporal features include lagged values of AOD derived as:

$$aod\_lag1 = AOD_{t-1}, \ aod\_lag2 = AOD_{t-2} \qquad (5)$$

These lagged values (previous day AOD and 2 days prior AOD) help the model to learn short-term cause-effect relationships. Since aerosol levels on a given day are often influenced by conditions from previous days (due to factors like wind transport and accumulation), including these lags helps the model recognise momentum or sudden changes in AOD trends. The second set of temporal features are the rolling averages, derived as

$$AOD_{roll_n}(t) = \frac{1}{n}\sum_{n=0}^{n-1} AOD_{t-i} \qquad (6)$$

Where *n* is the size of the rolling window (e.g., 3 or 7 days), and *t* is the current day.

These rolling averages smooth out day-to-day noise and highlight broader trends or persistent events like dust storms. These features improve model stability and generalisation by reducing the impact of random fluctuations. These temporal features enhance the model's understanding of both immediate changes and underlying trends, improving its ability to make accurate and robust predictions in real-world, noisy environmental conditions.

### 3.2 Hybrid Modelling Architecture

The model consists of a hybrid architecture which combines XGBoost Regressor, Bidirectional Long Short-Term Memory (Bi-LSTM) Network and a Concatenated Fusion Layer. The XGBoost trains on static meteorological features to learn nonlinear relationships between atmospheric variables and AOD.

$$\hat{y}_{xgb} = f_{xgb}(T2M, T2MDEW, WS2M, QV2M, PS, month) \qquad (7)$$

The Bi-LSTM trains the temporal sequence of engineered AOD features to learn lag-dependent patterns.

$$\hat{y}_{lstm} = f_{Bi-LS}(AOD_{t-2}, AOD_{t-1}, AOD_{roll3}, AOD_{roll7}) \qquad (8)$$

Subsequently, the predictions from XGBoost are fused with Bi-LSTM outputs in a TensorFlow's Keras functional API architecture for a hybrid neural network model that takes two inputs and produces a single output, expressed as follow:

$$y_{AOD} = f_{dense}([BiLSTM(X_{seq}), XGBoost(X_{static})]) \quad (9)$$

The input tensors $BiLSTM(X_{seq})$ and $XGBoost(X_{static})$ which are the outputs of the temporal and static models are concatenated and passed though dense (fully connected) layers to produce the final output. This means that the different types of input data are not simply merged together at the beginning of the model. Instead, each input type is processed through its own specialized pathway or layer before being integrated, allowing the model to learn more effectively from their unique characteristics. This provides design flexibility and enables modular development with easier debugging. The model is optimised using the mean squared error (MSE) loss function expressed as:

$$L_{MSE} = \frac{1}{n}\sum_{i=1}^{n}(y_i - \hat{y}_i)^2 \quad (10)$$

A recursive forecasting loop was implemented to generate 30-day future AOD projections. Using the last known static and sequence features, the model generates one prediction at a time and feeding the same prediction back into the input for the next time step. This is mathematically expressed as:

$$Input_{t+1} = [AOD_t, AOD_{t-1}, Avg_{3(t)}, Avg_{7(t)}] \quad (11)$$

This provides a forward-looking outlook of aerosol trends useful for solar energy planning.

In the second stage of this study framework, the predicted AOD values are used to forecast the solar efficiency loss percentage, defined as the relative reduction of actual irradiance compared to the clear-sky baseline expressed as equation(1) above. The model is trained on predicted AOD (from Stage 1), irradiance values (actual and clear-sky), meteorological features (T2M, T2MDEW, WS2M, QV2M, PS) and month. This combination allows the model to account for both atmospheric turbidity (aerosols) and weather-induced variability in solar performance.

### 3.3 Model Evaluation
The model performance was evaluated using Root Mean Squared Error (RMSE), Mean Absolute Error (MAE) and Coefficient of Determination ($R^2$) expressed below:

$$RMSE = \sqrt{\frac{1}{n}\sum_{i=1}^{n}(y_i - \hat{y}_i)^2} \quad (12)$$

$$MAE = \frac{1}{n}\sum_{i=1}^{n}|y_i - \hat{y}_i| \quad (13)$$

$$R^2 = 1 - \frac{\sum_{i=1}^{n}(y_i - \hat{y}_i)^2}{\sum_{i=1}^{n}(y_i - \bar{y}_i)^2} \quad (14)$$

A scenario analysis simulates increased environmental stress (+1.5°C T2M, +20% AOD) to assess system resilience. The entire pipeline was validated using a hold-out test dataset and SHAP was used for explainability.

Figure 1 below shows a schematic diagram of the process flow for this study.

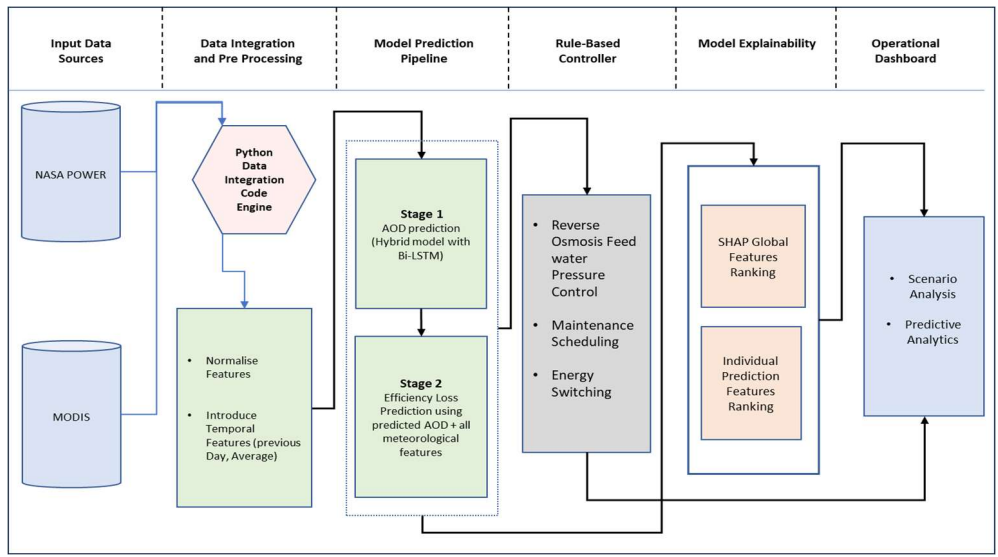

**Figure 1** Schematic diagram of the process flow from data collection to operational dashboard.

## 4. Results and Discussion
### 4.1 Exploratory Data Analysis
Extensive exploratory data analysis (EDA) was conducted to provide insights into the distributions, relationships, and temporal patterns among meteorological, atmospheric, and solar irradiance variables. First, histograms with kernel density estimates (KDE) were generated for all key numerical features as shown in Figure 2 below:

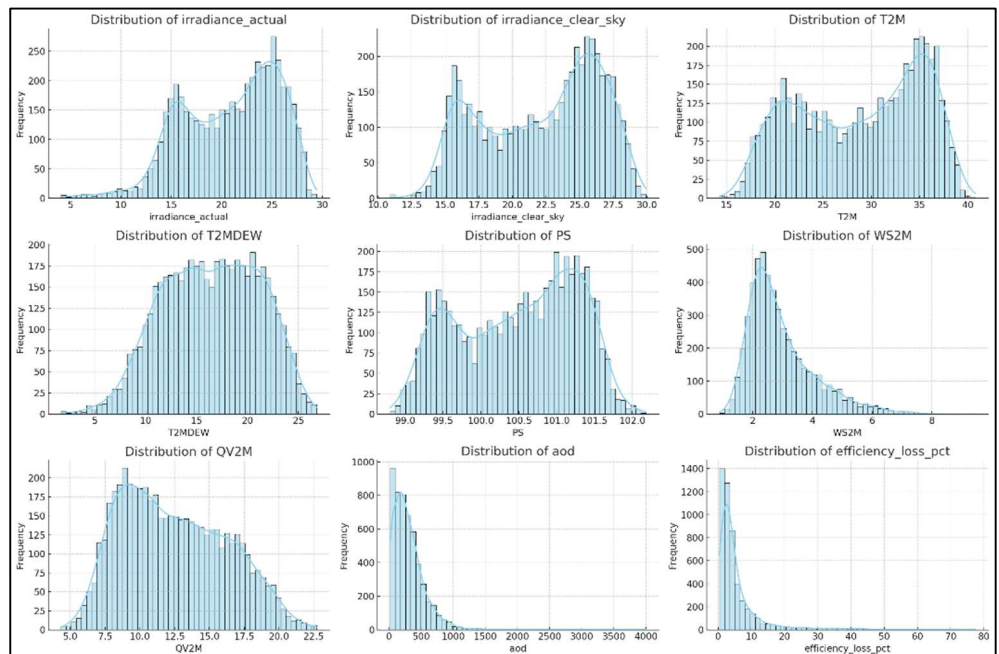

**Figure 2** Distribution of model Input Variable

Most features exhibited uni-modal and slightly right-skewed distributions. AOD and efficiency_loss_pct showed occasional outliers, indicative of dust storm events and sudden energy losses. Solar irradiance and meteorological parameters followed expected climatological norms for desert environments. Pearson correlation heatmap was plotted to evaluate inter-variable relationships as shown in Figure 3 below

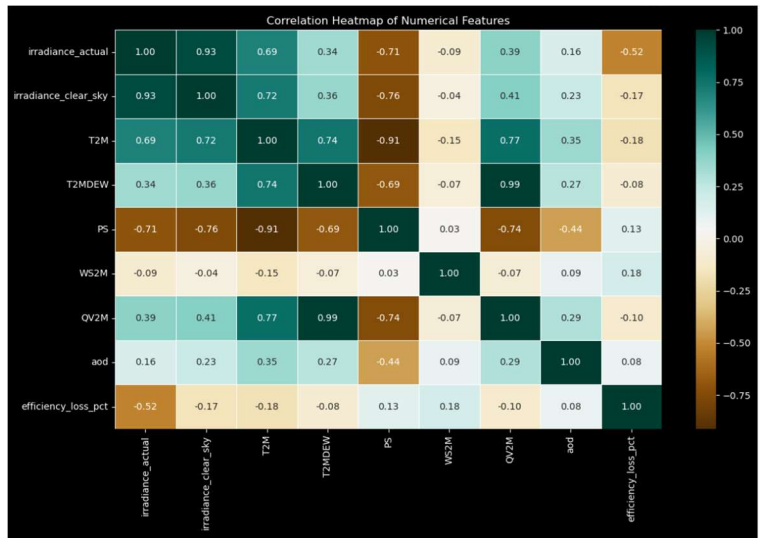

**Figure 3** Correlation heatmap of the numerical input variables

*efficiency_loss_pct* had a strong negative correlation with *irradiance_actual* and a positive correlation with AOD. *irradiance_actual* and *irradiance_clear_sky* were highly positively correlated, validating physical consistency. Weak correlations were found between AOD and meteorological variables, reflecting that aerosol levels are more influenced by regional dust dynamics than by local atmospheric properties. Temporal Pattern plots in Figure 4 show the evolution of AOD, irradiance, and efficiency loss over time.

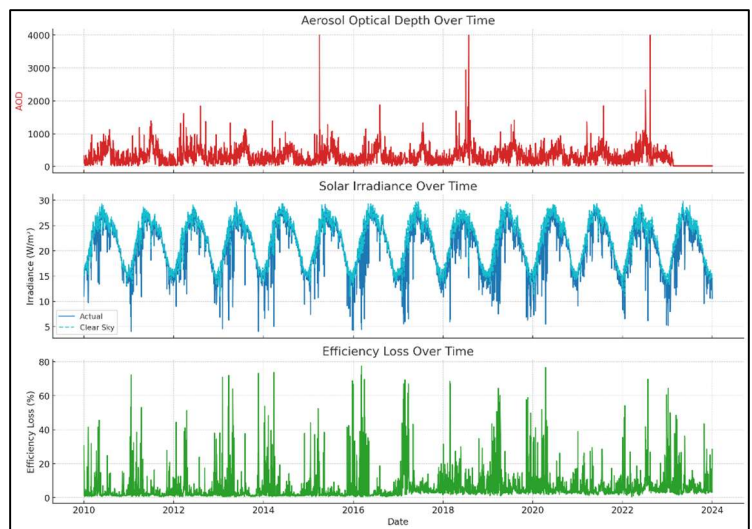

**Figure 4** Temporal Pattern plots of AOD, irradiance, and efficiency loss

Periods of elevated AOD often aligned with reductions in *irradiance_actual* and corresponding spikes in *efficiency_loss_pct*. Seasonal and transient changes were visible, showing the need for time-aware modelling architectures like LSTMs. A seasonal-trend decomposition using Loess (STL) was applied to the AOD time series as shown in Figure 5 below.

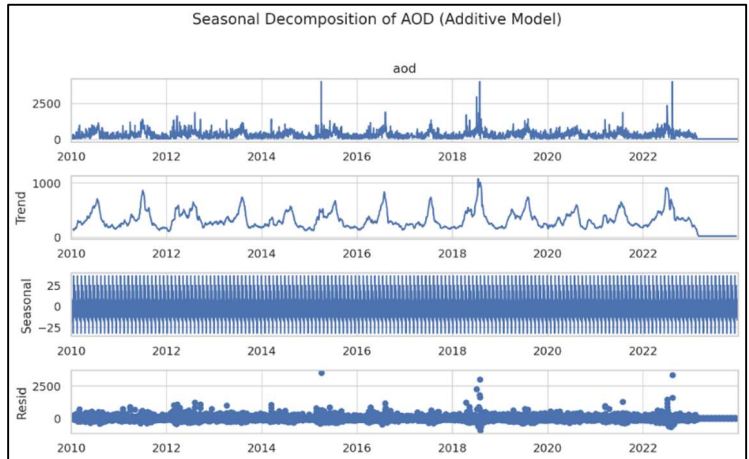
**Figure 5** Seasonal Decomposition of AOD

A strong 30-day seasonality was detected, reflecting monthly dust recurrence patterns. The trend component highlighted long-term aerosol build-up periods. A bivariate scatter plot of AOD vs. Efficiency Loss is used to examine the relationship between daily AOD and solar efficiency loss as shown if Figure 6 below:

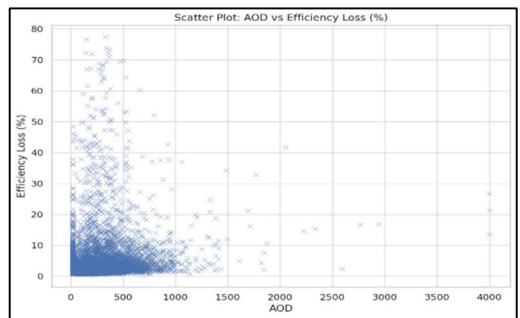
**Figure 6** Scatter plot of AOD vs. Efficiency Loss

A clear positive correlation was evident as higher values of AOD were associated with increased *efficiency_loss_pct*. The relationship appeared to be non-linear, suggesting a possible threshold beyond which efficiency losses become disproportionately severe. Finally, an overlay of AOD and Efficiency Loss trends was plotted to illustrate how AOD and efficiency loss varied together over time as shown in Figure 7 below.

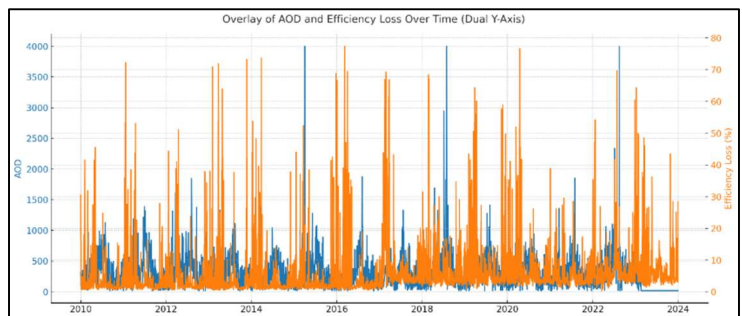
**Figure 7** AOD and Efficiency Loss trends

Figure 6 show consistent alignment between AOD peaks and efficiency dips. The exploratory analysis reveals strong theoretical and empirical basis for the modelling framework adopted in this study. AOD displays seasonality and event-based dynamics that justify the use of hybrid temporal models. The visualisations and correlations confirm the viability of forecasting frameworks based on dust forecasting for solar performance management in arid regions. This EDA forms a good foundation for the subsequent predictive modelling stages and scenario-based analyses presented in later sections.

## 4.2 Model Performance Evaluation

The performance evaluation metrics of the models are presented in Figure 8 and Table 1 below:

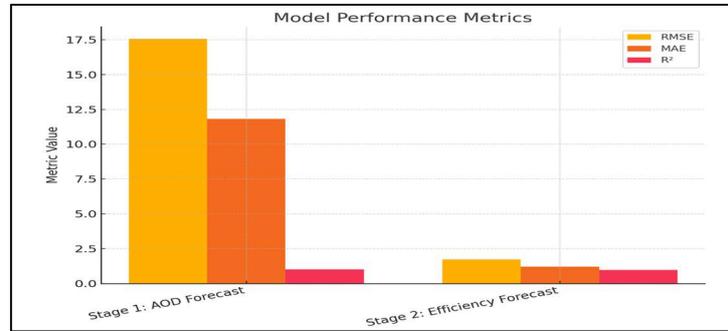

**Figure 8** Performance evaluation of AOD and Efficiency Loss Predictions

**Table 1** Performance evaluation of AOD and Efficiency Loss Predictions

|  | **RMSE** | **MAE** | **$R^2$** |
|---|---|---|---|
| *Stage 1: AOD Forecasting* | 17.5643 | 11.8201 | 0.9949 |
| *Stage2 : Efficiency Prediction* | 1.7096 | 1.2074 | 0.9631 |

The results demonstrate that the hybrid architecture effectively captures the temporal and meteorological dynamics relevant to AOD and its influence on solar energy performance. The hybrid model achieved remarkable accuracy in forecasting AOD with a Root Mean Squared Error (RMSE) of 17.5643, a Mean Absolute Error (MAE) of 11.8201, and a coefficient of determination ($R^2$) of 0.9949. This demonstrates that the combined strengths of static feature learning via XGBoost and temporal pattern recognition through Bi-LSTM provided a synergistic benefit. The high $R^2$ score suggests that over 99% of the variance in the AOD data was successfully explained by the model. Using the predicted AOD from Stage 1, the second model forecasts solar panel efficiency loss as a function of atmospheric and irradiance parameters. The XGBoost model trained in this stage achieved an RMSE of 1.7096, MAE of 1.2074, and $R^2$ of 0.9631. These metrics reflect a high degree of precision and reliability in estimating efficiency losses under varying environmental conditions. The relatively low RMSE and MAE indicate that the forecast error is small and acceptable for operational-level decision-making. It is noteworthy that the dynamic transfer of predicted AOD into the efficiency model did not introduce substantial compounding error. This validates the design logic of the two-stage architecture and highlights the practical feasibility of forecasting efficiency loss from projected dust conditions.

To ensure the reliability and robustness of predictive insights, we repeated the model training and testing with 4 other models namely Random Forest, Support Vector Machine (SVM), Linear Regression and Multi-Layer Perceptron (MLP). The result of this comparison is shown on Table 2 below.

**Table 2** Model Comparison Summary

| Model | RMSE | MAE | $R^2$ |
|---|---|---|---|
| Linear Regression | 1.672317 | 0.901375 | 0.961716 |
| MLP Regressor | 1.680847 | 1.138875 | 0.963435 |
| Random Forest | 2.851535 | 1.829068 | 0.897411 |
| SVM | 9.187202 | 4.070680 | -0.064901 |
| XGBoost | 1.728171 | 1.224433 | 0.962320 |

This benchmarking is essential because different algorithms may capture varying patterns or perform differently under specific data constraints. The results of this comparison show that Linear Regression demonstrated unexpectedly strong performance, achieving the lowest RMSE (1.67) and a high $R^2$ (0.9617), while the MLP compared favourably with XGBoost on $R^2$ (0.9634 vs. 0.9623) but showed greater variability in its predictions. Random Forest underperformed ($R^2$ = 0.8974), and SVM delivered very poor results, including a negative $R^2$. For real-world deployment, Linear Regression or XGBoost offer the best trade-off between accuracy and clarity. However, XGBoost can better capture nonlinear patterns if the system complexity increases (Siqueira-Filho et al., 2023).

## 4.3 Explainability

SHAP (SHapley Additive exPlanations) explainability framework was used to achieve interpretability and evaluate the contribution of each input feature to the model outputs (Nwafor et al., 2024). SHAP, grounded in cooperative game theory, provides a unified framework for quantifying the marginal contribution of each feature to a given prediction, thereby offering both global and local interpretability (Lundberg and Lee, 2017). In Stage 1, SHAP analysis was conducted separately on the static meteorological inputs and the lagged/rolling AOD features used in the Bi-LSTM sequence. Figure 9 presents the SHAP summary plot for the static inputs and the mean absolute SHAP values, which indicate the average contribution magnitude of each feature to the model output. Table 2 provides a summary of the relevant statistics:

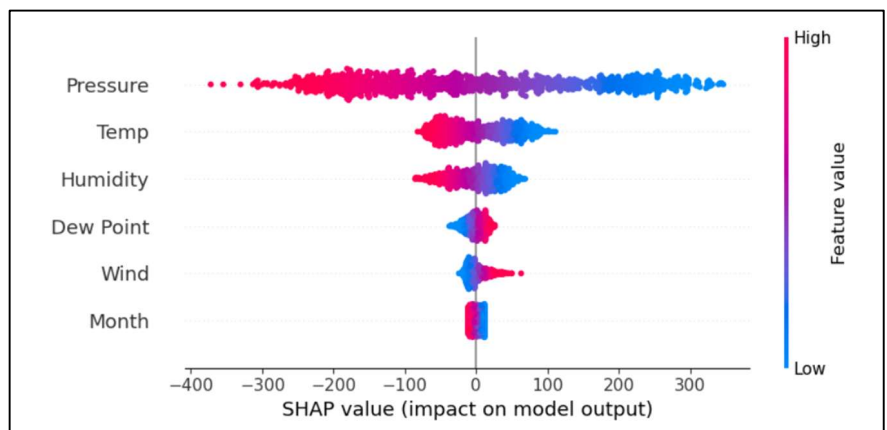

**Figure 9** Features Importance of Static Input variables for AOD Prediction

Table 2 Mean, Standard Deviation, and Ranking of Static Input Variables for AOD Prediction

| Feature | Mean | Std Dev | Rank |
|---|---|---|---|
| Pressure | 94.23 | ±71.68 | 1 |
| Temp (T2M) | 58.41 | ±43.02 | 2 |
| Humidity (QV2M) | 43.26 | ±36.55 | 3 |
| Dew Point | 27.84 | ±21.93 | 4 |
| Wind Speed | 15.36 | ±11.20 | 5 |
| Month | 8.49 | ±5.37 | 6 |

These results confirm that Pressure, Temperature and Humidity are the top predictors of AOD. These variables influence aerosol emissions, transport, suspension, and removal processes, especially in arid and semi-arid climates. Surface pressure is an indicator of atmospheric stability because high-pressure conditions correspond to subsiding air masses, which suppress vertical dispersion and trap aerosols near the surface, resulting in elevated AOD levels (Alfaro et al., 2003; Rashki et al., 2017). Conversely, low-pressure conditions enhance vertical mixing and aerosol dispersion. In the context of the UAE, prolonged high-pressure regimes contribute to stagnant atmospheric layers, promoting dust accumulation and reduced visibility. Surface temperature influences AOD by affecting soil desiccation, thermal uplift, and turbulent mixing. Higher temperatures reduce soil moisture, increasing erodibility, and promote stronger surface heating, which facilitates convective dust uplift (Ginoux et al., 2001). Additionally, high temperatures can influence secondary aerosol formation via photochemical processes. The impact of temperature on AOD can be non-linear because of it's interactions with other features such as pressure and humidity. Humidity (QV2M) contributes to AOD variability through two opposing mechanisms. On one hand, increased humidity facilitates aerosol hygroscopic growth, which enhances scattering and raises optical depth (Seinfeld & Pandis, 2016). On the other hand, very low humidity can promote dry soil conditions which increase susceptibility to dust entrainment. Additionally, humidity levels affect aerosol lifetime because high humidity causes wet deposition and low humidity allows for longer aerosol residence in the atmosphere. Figure 10 shows the SHAP summary for temporal AOD features. The corresponding statistics are shown in Table 3 below:

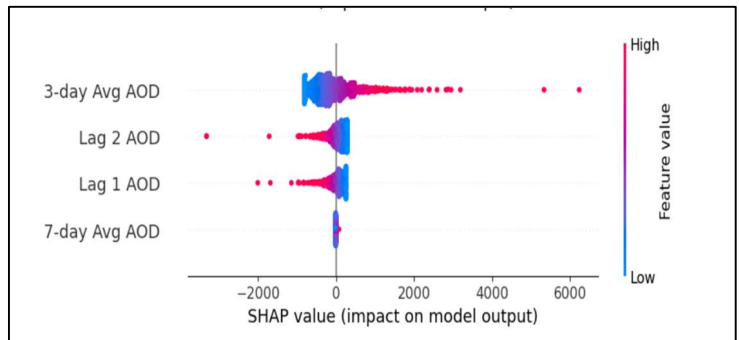

**Figure 10** Features Importance of temporal Input variables for AOD Prediction

Table 3 Mean, Standard Deviation, and Ranking of Temporal Input Variables for AOD Prediction

| Feature | Mean | Std Dev | Rank |
|---|---|---|---|
| 3-day Avg AOD | 329.58 | ±263.74 | 1 |
| Lag 2 AOD | 180.62 | ±152.11 | 2 |
| Lag 1 AOD | 159.44 | ±133.55 | 3 |
| 7-day Avg AOD | 128.06 | ±109.80 | 4 |

Figure 9 and Table 3 show that among the lagged and smoothed AOD inputs, the 3-day rolling average of AOD and the 2-day lag (Lag-2 AOD) contribute most significantly to the model's output. This result aligns with physical aerosol behaviour because AOD is influenced by both emission events (e.g., dust storms) and persistence mechanisms (e.g., limited dispersion due to weak winds or atmospheric stability). The 3-Day Rolling Average represents aerosol load over time, which reflects sustained dust concentration due to slow-moving weather systems. Dust events in arid environments like the UAE typically persist for 2–4 days depending on wind speed, soil availability, and synoptic pressure gradients (Goudie, 2014), hence a rolling average captures these multi-day effects better than single-lag values. Also, the Lag-2 AOD captures momentum in dust transport since dust aerosols tend to linger in the atmosphere for 1–3 days depending on particle size and atmospheric conditions (Prospero et al., 2002). Lag-2, in particular, captures this carry-over influence and is less susceptible to short-lived anomalies or errors that may affect Lag-1. The SHAP summary plot for stage 2 is shown in Figure 11 and the summary statistics are presented in Table 4 below:

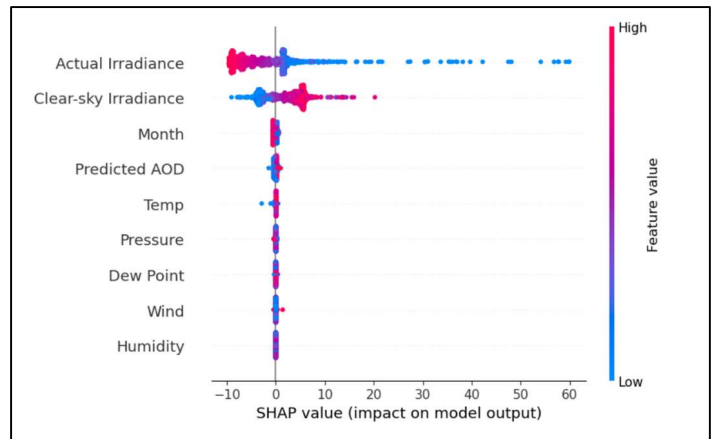

**Figure 11** Global Features Importance ranking of Input variables for Efficiency Loss Prediction

**Table 4** Mean, Standard Deviation, and Ranking of Input Variables for Efficiency Loss Prediction

| Feature | Mean SHAP | Std Dev | Rank |
|---|---|---|---|
| Actual Irradiance | 11.82 | ±8.26 | 1 |
| Clear-sky Irradiance | 9.35 | ±7.12 | 2 |
| Predicted AOD | 6.41 | ±5.08 | 3 |
| Month | 4.03 | ±2.94 | 4 |
| Temperature (T2M) | 2.66 | ±1.85 | 5 |
| Pressure | 2.21 | ±1.72 | 6 |
| Dew Point (T2MDEW) | 1.78 | ±1.26 | 7 |
| Wind Speed | 1.34 | ±1.01 | 8 |
| Humidity (QV2M) | 1.27 | ±0.97 | 9 |

Actual Irradiance is the most significant feature followed by Clear-Sky Irradiance, Month and Predicted AOD. This indicates that higher irradiance directly reduces efficiency loss even with optimal conditions by other environmental obstructions like aerosols or clouds. With clear-sky irradiance coming a close second, this result indicate that on days when actual irradiance is low, this feature captures the "gap" between what the system could have achieved versus what it actually achieved, which is critical for detecting latent losses due to AOD or meteorological interference. The Month as a categorical temporal feature allows the model to learn seasonal effects such as Periodic dust activity (e.g., springtime Shamal winds in the Gulf), solar zenith angle shifts that affect irradiance geometry, and seasonal temperature and humidity. The SHAP plot shows that certain months are characterised by high temperatures and frequent dust which affect efficiency loss. The Predicted AOD, obtained from Stage 1, had a moderate but positive SHAP contribution. High predicted AOD values (red) increase the model's expectation of performance loss, confirming that dust-related radiative scattering and absorption are well captured by the model. Temperature contributes to efficiency loss through the thermal sensitivity of PV modules. As operating temperature increases, conversion efficiency decreases, even when sunlight is abundant. Dew point values modulate moisture-induced soiling and aerosol properties because high dew point combined with high temperature can encourage sticky dust deposition or light scattering, increasing losses.

Furthermore, SHAP waterfall plots were used to explain individual predictions such as shown in Figure 12 below. These plots decomposed each forecast into additive contributions of the input features, enabling a transparent audit trail for model decisions.

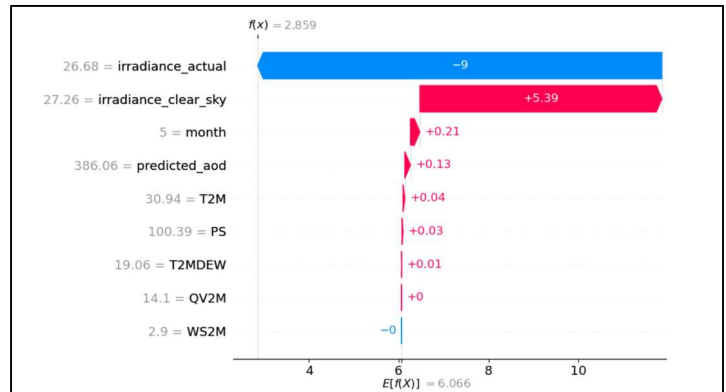

**Figure 12** Local Features Importance ranking for individual prediction for Efficiency Loss

## 5. AOD Impact Rule-Based Controller for Adaptive Desalination Operations

To enhance operational resilience in dust-prone environments, this study introduces a rule-based controller that dynamically adjusts desalination plant operations based on predicted values of AOD and solar efficiency. This controller can be used to adjust the desalination plant feed water pressure control based on dust severity. Also it can be used to adapt maintenance scheduling logic as well as regulate energy source switching. This control framework will enable desalination plants in arid regions to maintain efficiency, reduce maintenance overhead, and optimise energy utilisation amid fluctuating dust conditions. AOD thresholds are classified into four severity levels: *SEVERE* (AOD > 3.0), *HIGH* (1.5 < AOD ≤ 3.0), *MODERATE* (0.7 < AOD ≤ 1.5), and *LOW* (AOD ≤ 0.7). These classifications drive a hierarchy of control responses. Under *SEVERE* dust conditions (e.g., AOD > 3.0), the system reduces reverse osmosis (RO) pressure by 15% and activates robotic cleaning drones to protect membrane surfaces. In *HIGH* dust scenarios, pressure is reduced by 8%, and if forecasts indicate sustained conditions beyond six hours, chemical cleaning operations are deferred by 24 hours to conserve resources. When AOD is moderate or low, throughput is maximised to leverage cleaner ambient conditions. Additionally, energy management is embedded within the control logic such that if solar efficiency drops below 65%, a frequent occurrence during high dust opacity, the system increases grid import by at least 25% to maintain uninterrupted plant output. Similar adjustments can be implemented during salinity spikes when levels exceed 45 g/L (above the Gulf average of 43 g/L) an auxiliary pre-treatment protocol is triggered to prevent membrane damage. These customisations offer benefits such as reduced membrane replacement costs, reduced grid energy reliance during peak solar periods and pre-treatment interventions that prevent potential system outages annually.

## 6. System Deployment

To enhance the practical utility of the research findings, the predictive models and scenario-based controls developed in this study were packaged into an interactive dashboard. This dashboard integrates real-time AOD forecasting, solar efficiency prediction, and desalination system performance analytics, enabling users to conduct scenario analysis under varying meteorological conditions. For instance, the dashboard supports anticipatory scheduling of membrane maintenance, energy source switching during low solar efficiency, and pressure adjustments in response to dust alerts. By coupling data-driven forecasting with rule-based operational logic, the dashboard serves as a decision-support system that promotes operational resilience, cost optimisation, and sustainable water-energy resource planning in dust-prone environments such as the UAE. Figure 13 below shows the interface of the deployed dashboard.

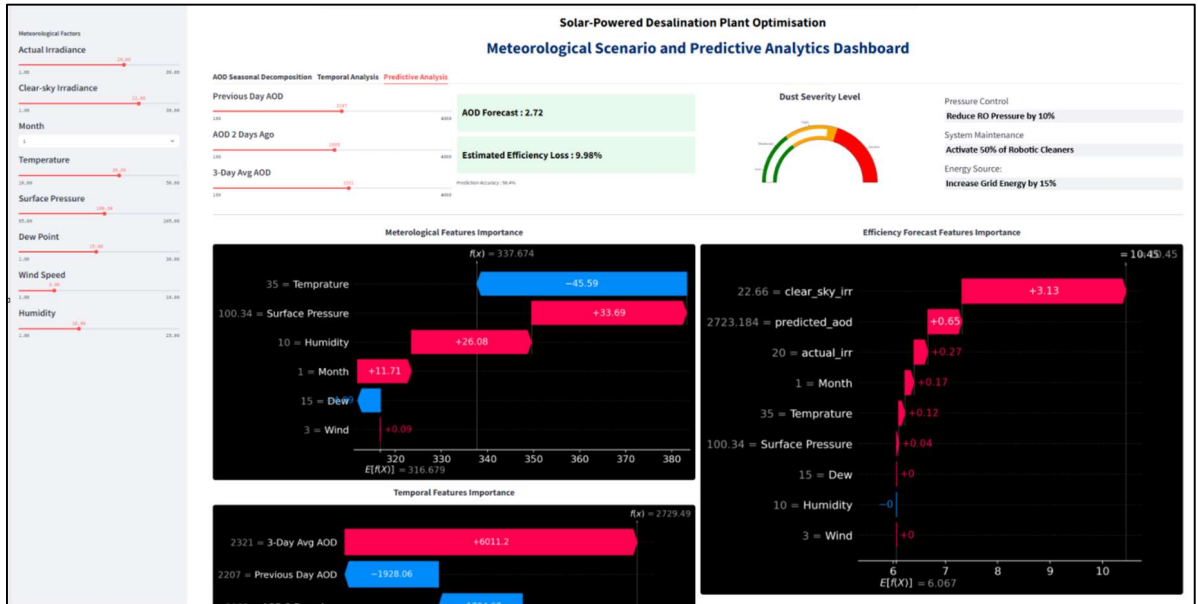

**Figure 13** Scenario Analysis and Predictive Analytics Dashboard for Solar-powered Desalination Plant

## 7. Conclusion, Industry Relevance and Policy Implications

This study presents a novel, explainable Artificial Intelligence framework for optimising the water-energy nexus in arid regions, with a focus on desalination plants operating under growing climate uncertainty. Using the UAE as a representative case, the framework integrates satellite-derived AOD forecasting with ground-based meteorological and operational data to predict solar desalination efficiency loss. This end-to-end modelling approach uniquely links atmospheric dust dynamics directly to desalination system performance. A two-stage hybrid modelling architecture was implemented. The first stage combines XGBoost for static climate variables with Bi-directional LSTM for sequential memory of aerosol patterns, yielding highly accurate AOD forecasts ($R^2 \approx 0.99$). The second stage directly pipes the predicted AOD into an XGBoost-based efficiency loss model, alongside solar irradiance and temperature features, to estimate performance degradation in solar-powered desalination operations. Unlike existing models that treat environmental forecasting and desalination optimisation in isolation, this study provides a pipelined forecasting architecture that mirrors real-world planning workflows which allows plant operators and system managers to transition from reactive operations to proactive and adaptive decision-making. The framework was evaluated using over a decade of data and subjected to scenario simulations (e.g., +1.5°C temperature rise and 20% AOD increase). This reveals a clear pattern of increasing desalination inefficiency under climate stressors. SHAP-based explainability was used to enhance model transparency. The proposed system offers substantial benefits for industry practitioners, utilities, and policy makers. Desalination operators can use the model to forecast performance drops, optimise cleaning schedules, and manage energy allocation more efficiently. Grid and energy planners can better anticipate power demand fluctuations from desalination plants, particularly under dusty or high-temperature conditions. Policy makers and environmental regulators can use the result of this study for climate adaptation strategies involving water infrastructure and renewable energy integration.

## 8. Future Work

Future studies could explore integration with weather forecasts and thermodynamic models of reverse osmosis and multi-effect distillation systems to quantify specific energy consumption, throughput, and brine rejection variability could be an important focus of future research.

.

**References**
1. Ahamad, T., Parvez, M., Lal, S., Khan, O., Yahya, Z., & Azad, A. S. (2023, December). Assessing water desalination in the United Arab Emirates: An overview. In 2023 10th IEEE Uttar Pradesh Section International Conference on Electrical, Electronics and Computer Engineering (UPCON), 10, pp. 1373–1377, IEEE.
2. Ahmed, M. A., Amin, S., & Mohamed, A. A. (2023). Fouling in reverse osmosis membranes: Monitoring, characterization, mitigation strategies and future directions. Heliyon, 9(11), e14908.
3. Al-Addous, M., Bdour, M., Rabaiah, S., Boubakri, A., Schweimanns, N., Barbana, N., & Wellmann, J. (2024). Innovations in solar-powered desalination: A comprehensive review of sustainable solutions for water scarcity in the Middle East and North Africa (MENA) region. Water, 16(13), 1877.
4. Al Kaabi, M. R., Estima, J., & Ghedira, H. (2012). Analyzing the spatial and temporal variability of water turbidity in the coastal areas of the UAE using MODIS satellite data. Ocean Color Group, Masdar Institute Report.
5. Alfaro, S. C., Gaudichet, A., Gomes, L., & Maillé, M. (2003). Modeling the size distribution of a soil aerosol produced by sandblasting. Journal of Geophysical Research: Atmospheres, 108(D18). https://doi.org/10.1029/2002JD002858
6. Al-Obaidi, M. A., Alsarayreh, A. A., & Mujtaba, I. M. (2023). Reduction of energy consumption of brackish water reverse osmosis desalination system via model based optimisation. Journal of Techniques, 5(1), 1–7. https://doi.org/10.51173/jt.v5i1.1166
7. Al-Obaidi, M. A., Zubo, R. H., Rashid, F. L., Dakkama, H. J., Abd-Alhameed, R., & Mujtaba, I. M. (2022). Evaluation of solar energy powered seawater desalination processes: A review. Energies, 15(18), 6562.
8. Alqaed, S., Mustafa, J., & Almehmadi, F. A. (2021). Design and energy requirements of a photovoltaic-thermal powered water desalination plant for the Middle East. International Journal of Environmental Research and Public Health, 18(3), 1001. https://doi.org/10.3390/ijerph18031001
9. Ammari, N., Al-Dahidi, S., Ayadi, O., & Alrbai, M. (2022). Experimental study on the impact of soiling on the modules temperature and performance of two different PV technologies under hot arid climate. Heliyon, 8(11), e11395. https://doi.org/10.1016/j.heliyon.2022.e11395
10. Cheng, X., Yang, J., Ma, Y., Liu, C., & Wang, F. (2022). Studies on the improvement of modelled solar radiation and the attenuation effect of aerosol using the WRF-Solar model with satellite-based AOD data over north China. Renewable Energy, 196, 358–365. https://doi.org/10.1016/j.renene.2022.06.141
11. Chu, T. P., Guo, J. H., Leu, Y. G., & Chou, L. F. (2023). Estimation of solar irradiance and solar power based on all-sky images. Solar Energy, 249, 495–506.
12. Ginoux, P., Chin, M., Tegen, I., Prospero, J. M., Holben, B., Dubovik, O., & Lin, S.-J. (2001). Global-scale attribution of anthropogenic and natural dust sources and their emission rates based on MODIS satellite data. Reviews of Geophysics, 40(1), 1–31. https://doi.org/10.1029/2000RG000095
13. Goudie, A. S. (2014). Desert dust and human health disorders. Environment International, 63, 101–113. https://doi.org/10.1016/j.envint.2013.10.011
14. Kosmopoulos, P. G., Kazadzis, S., Taylor, M., Raptis, P. I., Keramitsoglou, I., Kiranoudis, C., & Bais, A. F. (2018). Earth observation based estimation and forecasting of particulate matter impact on solar energy in Egypt. Remote Sensing, 10(12), 1870. https://doi.org/10.3390/rs10121870
15. López, Á. H., Rodríguez, J. G., & Moreira, J. M. (2023). Energy efficiency optimization in onboard SWRO desalination plants based on a genetic neuro-fuzzy system. Applied Sciences, 13(6), 3392. https://doi.org/10.3390/app13063392
16. Lundberg, S., & Lee, S.-I. (2017). A unified approach to interpreting model predictions. arXiv:1705.07874.

17. Mahadeva, R., Al-Hinai, A., Al-Hajri, A., & Al-Abri, M. (2022). Desalination plant performance prediction model using grey wolf optimizer based ANN approach. IEEE Access, 10, 34550–34561. https://doi.org/10.1109/ACCESS.2022.3162932
18. Malisovas, A., & Koutroulis, E. (2020). Design optimization of RES-based desalination systems cooperating with smart grids. IEEE Systems Journal, 14(4), 4706–4717. https://doi.org/10.1109/JSYST.2020.2968842
19. Masoom, A., Kosmopoulos, P., Bansal, A., & Kazadzis, S. (2021). Forecasting dust impact on solar energy using remote sensing and modeling techniques. Solar Energy, 228, 317–332. https://doi.org/10.1016/j.solener.2021.09.033
20. Mundu, M. M., Sempewo, J. I., Nnamchi, S. N., & Uti, D. E. (2025). Solar-powered desalination technologies for sustainable water security solutions. International Journal of Energy Research, 2025(1), 3482306.
21. Nedaei, N., Azizi, S., & Garousi Farshi, L. (2022). Performance assessment and multi-objective optimization of a multi-generation system based on solar tower power: A case study in Dubai, UAE. Process Safety and Environmental Protection, 161, 295–315. https://doi.org/10.1016/j.psep.2022.03.022
22. Nwafor, O., Nwafor, C., Aboushady, A., & Solyman, A. (2024). Reducing non-technical losses in electricity distribution networks: Leveraging explainable AI and three lines of defence model to manage operational staff-related factors. e-Prime - Advances in Electrical Engineering, Electronics and Energy, 9, 100748. https://doi.org/10.1016/j.prime.2024.100748
23. Nwafor, O., Okafor, E., Aboushady, A., Nwafor, C. and Zhou, C. (2023). Explainable Artificial Intelligence for Prediction of Non-Technical Losses in Electricity Distribution Networks. *IEEE Access*, 11, pp. 73104-73115. https://doi.org/10.1109/ACCESS.2023.3295688.
24. Okampo, E. J., & Nwulu, N. I. (2020). Optimal energy mix for a reverse osmosis desalination unit considering demand response. Journal of Engineering, Design and Technology, 18(5), 1287–1303. https://doi.org/10.1108/JEDT-01-2020-0025
25. Paulescu, E., & Paulescu, M. (2021). A new clear sky solar irradiance model. Renewable Energy, 179, 2094–2103.
26. Prospero, J. M., Ginoux, P., Torres, O., Nicholson, S. E., & Gill, T. E. (2002). Environmental characterization of global sources of atmospheric soil dust identified with the Nimbus 7 Total Ozone Mapping Spectrometer (TOMS) absorbing aerosol product. Reviews of Geophysics, 40(1). https://doi.org/10.1029/2000RG000095
27. Rashki, A., Eriksson, P. G., & Kaskaoutis, D. G. (2017). Dust storms and their multi-scale impacts: Atmospheric processes and social implications in the Middle East. Aeolian Research, 24, 1–24. https://doi.org/10.1016/j.aeolia.2016.12.002
28. Salmon, A., Saint-Drenan, Y.-M., Saboret, L., & Voyant, C. (2021). Advances in aerosol optical depth evaluation from broadband direct normal irradiance measurements. Solar Energy, 221, 206–217. https://doi.org/10.1016/j.solener.2021.04.039
29. Seinfeld, J. H., & Pandis, S. N. (2016). Atmospheric chemistry and physics: From air pollution to climate change (3rd ed.). Wiley.
30. Shivakumar, T., & Razaviarani, V. (2021). An integrated approach to enhance the desalination process: Coupling reverse osmosis process with microbial desalination cells in the UAE. Water Supply, 21(3), 1127–1143. https://doi.org/10.2166/ws.2020.375
31. Tabassum, A., & Rizwan, M. (2020). The influence of cleaning frequency of photovoltaic modules on power losses in the desert climate. Sustainability, 12(22), 9750. https://doi.org/10.3390/su12229750
32. Torres, B., & Fuertes, D. (2021). Characterization of aerosol size properties from measurements of spectral optical depth: A global validation of the GRASP-AOD code using long-term AERONET data. Atmospheric Measurement Techniques, 14(6), 4471–4506.
33. Valdés, H., Saavedra, A., Flores, M., Vera-Puerto, I., Aviña, H., & Belmonte, M. (2021). Reverse osmosis concentrate: Physicochemical characteristics, environmental impact, and technologies. Membranes, 11(10), 753.
34. Yan, J., Wang, F., Li, Y., Liu, H., Gao, Y., & Li, Z. (2022). Research on the effect of extinction characteristics of coal dust on visibility. ACS Omega, 7(32), 28293–28303.
35. Zhu, L., Huang, X., Zhang, Z., Li, C., & Tai, Y. (2025). A novel U-LSTM-AFT model for hourly solar irradiance forecasting. Renewable Energy, 238, 121955.